\pdfoutput=1

\documentclass[11pt]{article}

\usepackage[preprint]{acl}

\usepackage{times}
\usepackage{latexsym}

\usepackage[T1]{fontenc}

\usepackage[utf8]{inputenc}

\usepackage{microtype}

\usepackage{inconsolata}
\usepackage{booktabs}
\usepackage{adjustbox}
\usepackage{tabularx}
\usepackage{multirow}
\usepackage{enumitem}
\usepackage{array}
\usepackage{caption}
\usepackage{hyperref}

\usepackage{graphicx}
\usepackage{subcaption}
\usepackage{amsmath}

\title{Beyond Human-Only: Evaluating Human-Machine Collaboration for Collecting High-Quality Translation Data}

\author{Zhongtao Liu, Parker Riley,  Daniel Deutsch, Alison Lui, \\ {\bf Mengmeng Niu, Apu Shah and Markus Freitag} \\
Google
\\
\texttt{\{zhongtao,prkriley,dandeutsch,alisonlui,mniu,apurva,freitag\}@google.com}
}

\newcolumntype{L}[1]{>{\raggedright\let\newline\\\arraybackslash\hspace{0pt}}m{#1}}
\newcolumntype{C}[1]{>{\centering\let\newline\\\arraybackslash\hspace{0pt}}m{#1}}
\newcolumntype{R}[1]{>{\raggedleft\let\newline\\\arraybackslash\hspace{0pt}}m{#1}}

\begin{document}
\maketitle
\begin{abstract}

Collecting high-quality translations is crucial for the development and evaluation of machine translation systems. However, traditional human-only approaches are costly and slow. This study presents a comprehensive investigation of 11 approaches for acquiring translation data, including human-only, machine-only, and hybrid approaches. Our findings demonstrate that human-machine collaboration can match or even exceed the quality of human-only translations, while being more cost-efficient. Error analysis reveals the complementary strengths between human and machine contributions, highlighting the effectiveness of collaborative methods. Cost analysis further demonstrates the economic benefits of human-machine collaboration methods, with some approaches achieving top-tier quality at around 60\% of the cost of traditional methods. We release a publicly available dataset\footnote{The dataset can be found at \url{https://github.com/google-research/google-research/tree/master/collaborative-tr-collection}.} containing nearly 18,000 segments of varying translation quality with corresponding human ratings to facilitate future research.

\end{abstract}

\section{Introduction}
\label{sec:intro}
Collecting high-quality translations efficiently presents significant challenges. Traditional approaches rely heavily on different tiers of human translators, ranging from professional linguists to junior bilingual speakers~\citep{zouhar2024quality}. While these approaches can produce high-quality translations, they are often expensive, time-consuming, and challenging to scale for large datasets.

Recent advancements in machine translation with large language models ~\citep{openai2024gpt4,geminiteam2024gemini} have demonstrated models' impressive abilities to generate human-like translations. However, recent research~\citep{yan2024gpt4vshumantranslators} tends to position human translators and machine translation systems as competitors rather than potential collaborators, which could result in efficient alternatives for addressing the limitations of traditional translation data collection methods. 

\begin{figure}[t]
  \includegraphics[width=\columnwidth]{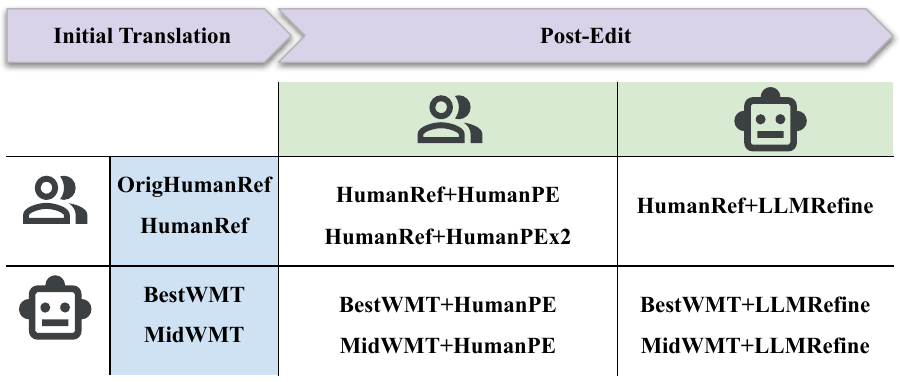}
  \caption{Our 11 translation systems, organized by initial translation type (human or machine) and post-editing type (none, human, or machine). Detailed system descriptions are provided in Section~\ref{sec:system_description}.
}
  \label{fig:annotation_types}
\end{figure}

In this paper, we aim to fill the gap by comprehensively investigating the potential of human-machine collaboration to efficiently collect high-quality translation data. We hypothesize that combining the strengths of humans and machines could lead to higher quality, cost-efficient translation collection methods. To verify the hypothesis, we explore 11 different methods for acquiring translation data, including human-only, machine-only, and various hybrid methods.

Our research seeks to answer the following key questions:

\begin{itemize}
    \item Can human-machine collaborative approaches produce translations of comparable or higher quality than traditional human-only or machine-only methods?
    \item How do different collaborative methods impact translation quality, and where do the improvements primarily originate? 
    \item What are the cost implications of these various approaches, and can human-machine collaboration offer a more cost-efficient solution for high-quality translation collection?

\end{itemize}

Our findings demonstrate that human-machine collaboration can match or even exceed human-only translation quality while being more cost-efficient. We present detailed error analyses to reveal the complementary strengths of the collaborative methods and conduct a thorough cost analysis to illustrate the economic benefits of collaborative approaches.

To support future research, we also release a publicly available dataset containing nearly 18,000 segments of varying translation quality with corresponding human ratings. 


\section{Collecting Translations}
\label{sec:experiment_setup}

Translating text from one language to another can either be done by bilingual annotators or machine translation systems.
However, both cases are prone to producing errors in their translations, including well-trained expert translators \citep{freitag-etal-2023-results}.
As such, translations can be post-edited, a process of correcting a translation, either manually or with a model, that often yields higher-quality translations.

Both steps of this process --- the initial translation collection and the post-editing --- can either be done with humans or with models, each with their own advantages and disadvantages in terms of speed, quality, cost, and scalability.
In this work, we explore how combinations of human and machines for both steps of this pipeline can combine to produce high-quality translations.

\subsection{Data Sources}

We use the test sets provided by the WMT23 General MT Shared Task ~\citep{kocmi-etal-2023-findings} and collect new translations using several methods. These data sets comprise 460 English-German (EnDe) paragraph-level segments and 1175 Chinese-English (ZhEn) sentence-level segments with human rating annotations.

\subsection{Data Collection Systems}
\label{sec:system_description}

Figure~\ref{fig:annotation_types} illustrates the combinations from the two dimensions: initial translation and post-editing methods from either human annotators or machines. Machines may be either large language models (LLMs) or machine translation (MT) systems. This results in the 11 systems in the figure, named according to the source of the initial translation with a suffix representing the post-editing approach.

In this work, we use several different sources for the initial translation:

\begin{itemize}
    \item \textbf{OrigHumanRef} and \textbf{HumanRef} are human translations collected by professional translators. We refer to the original reference provided by the WMT23 General MT Shared Task \citep{kocmi-etal-2023-findings} as \textsc{OrigHumanRef}. We collected a new from-scratch professional translation \textsc{HumanRef} following the standard annotation steps.
    \item \textbf{BestWMT} is the top-ranked MT system picked from the official results of WMT23 General Translation Task: GPT4-5shot ~\citep{openai2024gpt4} for EnDe and Lan-BridgeMT~\citep{wu-hu-2023-exploring} for ZhEn, representing the state-of-the-art MT capability we can access.
    \item \textbf{MidWMT} is a middle-ranked MT system from the official results of WMT23 General Translation Task: ONLINE-G for both EnDe and ZhEn, representing the conventional MT quality we can use.
\end{itemize}

We additionally explore the following different methods for post-editing translations:

\begin{itemize}
    \item \textbf{HumanPE} refers to the post-edit service provided by a separate batch of linguists. \textbf{HumanPEx2} means the translation going through two independent rounds of post-edits from professional translators. There is no translator overlap between the two batches.
    \item \textbf{LLMRefine}~\citep{xu2024llmrefine} is one of the state-of-the-art post-edit approaches leveraging error feedback for pin-pointing corrections. Here we reproduced its error-feedback process and leverage Gemini-1.0 Ultra~\cite{geminiteam2024gemini} with the reported prompts to generate post-edited text.
\end{itemize}

\subsection{Evaluation}

In this paper, we use Multidimensional Quality Metrics \citep[MQM; ][]{lommel2014mqm,freitag-etal-2021-experts} to evaluate translation quality. MQM is the state-of-the-art human evaluation framework for MT. In MQM, expert raters identify error spans within translations, which are automatically converted to numeric scores. Lower scores indicate fewer errors and thus higher quality.

\section{Data Quality Analysis}
\label{sec:quality_analysis}
In this section, we seek to understand how our collected translations differ from each other (\S\ref{sec:lexical_overlap}), how those differences correspond to changes in quality (\S\ref{sec:human_eval}), and what those results indicate about the value of human-machine collaboration in terms of quality (\S\ref{sec:human_machine_collaboration}).

\subsection{Lexical Overlap and Similarity}
\label{sec:lexical_overlap}


\begin{figure}[t]
  \includegraphics[width=\columnwidth]{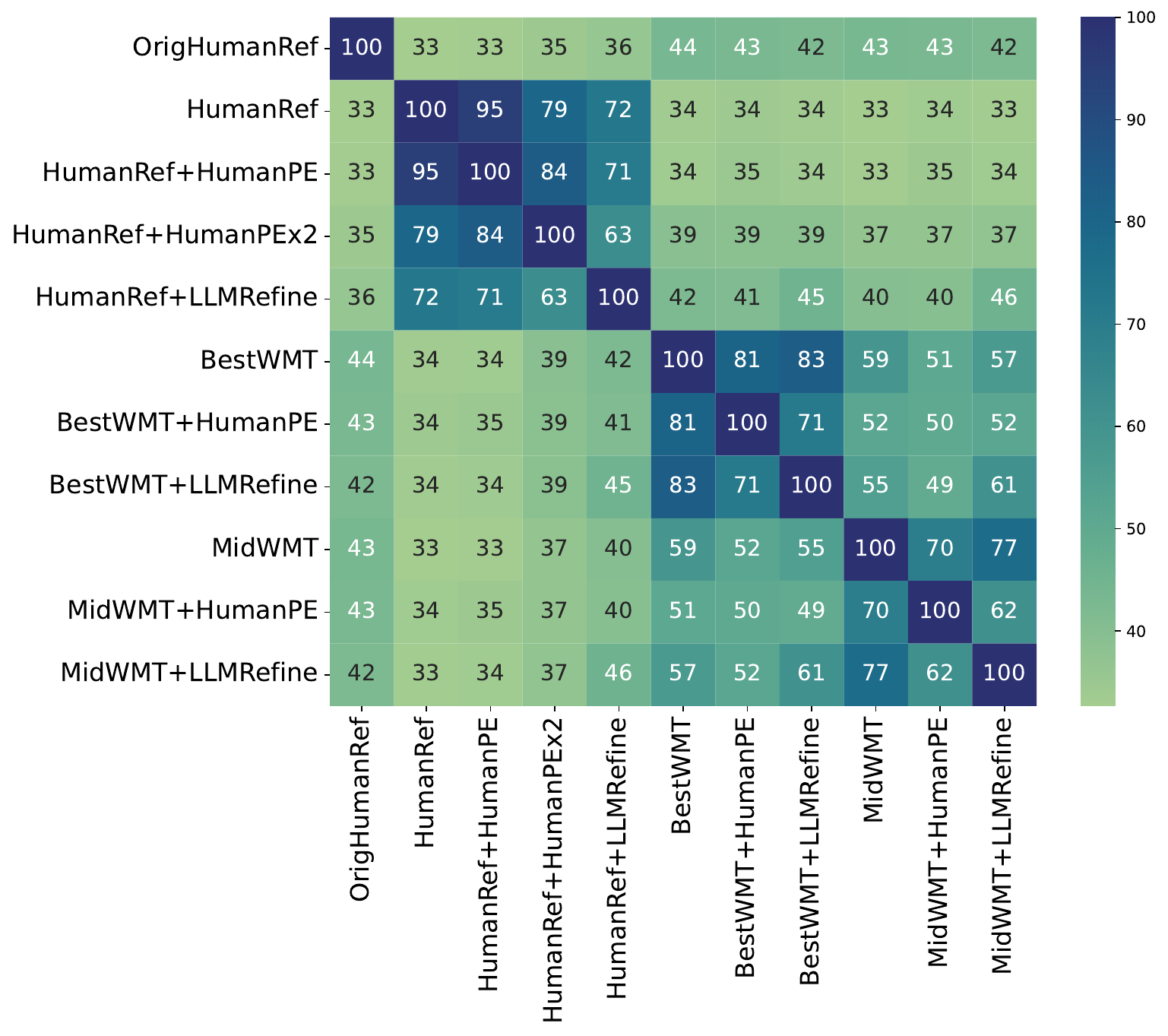}
  \caption{Cross-BLEU scores for different EnDe translation collection approaches.}
  \label{fig:cross_bleu}
\end{figure}

Figure ~\ref{fig:cross_bleu} presents a cross-BLEU~\citep{freitag-etal-2022-high} similarity matrix for English-German translations, which measures lexical similarity between pairs of translations. See Figure~\ref{fig:cross_bleu_zhen} in Appendix~\ref{sec:cross_bleu_appendix} for Chinese-English results. Higher scores indicate greater similarity.

One prominent pattern in these results is that systems based on the same initial translation retain high similarity even after post-editing. This indicates that \textbf{post-editing still preserves some characteristics of the original translation}. Also, translation systems based on MT (\textsc{BestWMT}, \textsc{MidWMT}, and their post-edited versions) are more similar to each other than to translations based on an initial human translation.


\begin{table}[t]
    \centering
    \begin{adjustbox}{width=0.98\columnwidth}
    \begin{tabular}{L{2.5cm}C{3.2cm}C{3.2cm}}
        \toprule
        \textbf{Source} & \bf \textsc{+ HumanPE}   & \bf \textsc{ + LLMRefine}\\
        \midrule
        \textsc{En-De} &  & \\
        \midrule
        \textsc{HumanRef} & 95 & \bf 72\\
        \textsc{BestWMT} & \bf 81 & 83\\
        \textsc{MidWMT} & \bf 70 & 77\\
        \midrule
        \textsc{Zh-En}  & &\\
        \midrule
        \textsc{HumanRef} & 88 & \bf 79\\
        \textsc{BestWMT} & \bf 84 & 89\\
        \textsc{MidWMT} & \bf 68 & 71\\
        \bottomrule
    \end{tabular}
    \end{adjustbox}
    \caption{Cross-BLEU score comparison between different post-edited versions of the same translation. Lower numbers indicate less similarity and more changes from the initial translation.}
    \label{tab:post-edit-cross-blue-compare}
\end{table}

Table~\ref{tab:post-edit-cross-blue-compare} presents a subset of the information in Figures~\ref{fig:cross_bleu} and ~\ref{fig:cross_bleu_zhen}, to emphasize the interaction between using human- vs. model-based approaches for the initial translation and post-edit. This illustrates the trend that \textbf{humans and machines tend to make more changes to translations from the other group}.

\subsection{MQM Quality Evaluation}
\label{sec:human_eval}

Tables ~\ref{tab:ende_mqm} and ~\ref{tab:zhen_mqm} present the MQM human evaluation results. The solid lines denote \textit{significance clusters}, where every system in a cluster is statistically significantly better than every system below that cluster, based on random permutation tests with 10,000 trials, where a $p$-value of less than $\alpha = 0.05$ is considered significant.

\begin{table}[t]
    \centering
    \begin{adjustbox}{width=0.98\columnwidth}
    \begin{tabular}{lc}
        \toprule
        \textbf{Reference} & \textbf{MQM per segment}\\
        \midrule
        \textsc{HumanRef+LLMRefine} & 2.76\\
        \textsc{OrigHumanRef} & 2.80\\
        \textsc{BestWMT+HumanPE} & 2.98\\
        \midrule
        \textsc{BestWMT} & 3.30\\
        \textsc{BestWMT+LLMRefine} & 3.36\\
        \textsc{HumanRef+HumanPE} & 3.53\\
        \textsc{HumanRef+HumanPEx2} & 3.70\\
        \midrule
        \textsc{HumanRef} & 3.79\\
        \textsc{MidWMT+HumanPE} & 3.83\\
        \textsc{MidWMT+LLMRefine} & 4.02\\
        \midrule
        \textsc{MidWMT} & 6.45\\
        \bottomrule
    \end{tabular}
    \end{adjustbox}
    \caption{English-German MQM human evaluation results. Lower scores represent higher translation quality.}
    \label{tab:ende_mqm}
\end{table}

\begin{table}[t]
    \centering
    \begin{adjustbox}{width=0.98\columnwidth}
    \begin{tabular}{lc}
        \toprule
        \textbf{Reference} & \textbf{MQM per segment}\\
        \midrule
        \textsc{HumanRef+LLMRefine} & 1.82\\
        \textsc{HumanRef+HumanPEx2} & 1.82\\
        \textsc{BestWMT+HumanPE} & 1.87\\
        \textsc{HumanRef+HumanPE} & 1.91\\
        \textsc{BestWMT+LLMRefine} & 1.94\\
        \midrule
        \textsc{HumanRef} & 2.05\\
        \textsc{BestWMT} & 2.22\\
        \textsc{MidWMT+HumanPE} & 2.23\\
        \midrule
        \textsc{MidWMT+LLMRefine} & 2.45\\
        \midrule
        \textsc{MidWMT} & 3.98\\
        \midrule
        \textsc{OrigHumanRef} & 5.63\\
        \bottomrule
    \end{tabular}
    \end{adjustbox}
    \caption{Chinese-English MQM human evaluation results. Lower scores represent higher translation quality.}
    \label{tab:zhen_mqm}
\end{table}

The results reveal that \textsc{\bf HumanRef+LLMRefine} and \textsc{\bf BestWMT+HumanPE} are the overall winners, with each appearing in the best significance cluster in both language pairs.

Tables~\ref{tab:ende_mqm} and \ref{tab:zhen_mqm} show that post-edits, both \textsc{HumanPE} and \textsc{LLMRefine}, \textbf{demonstrate a positive impact on initial translations}. These methods consistently either elevate the translation quality to a higher level of significance or preserve the existing quality.

\subsection{Human-Machine Collaboration}
\label{sec:human_machine_collaboration}


\begin{figure}[t]
  \includegraphics[width=0.95\columnwidth]{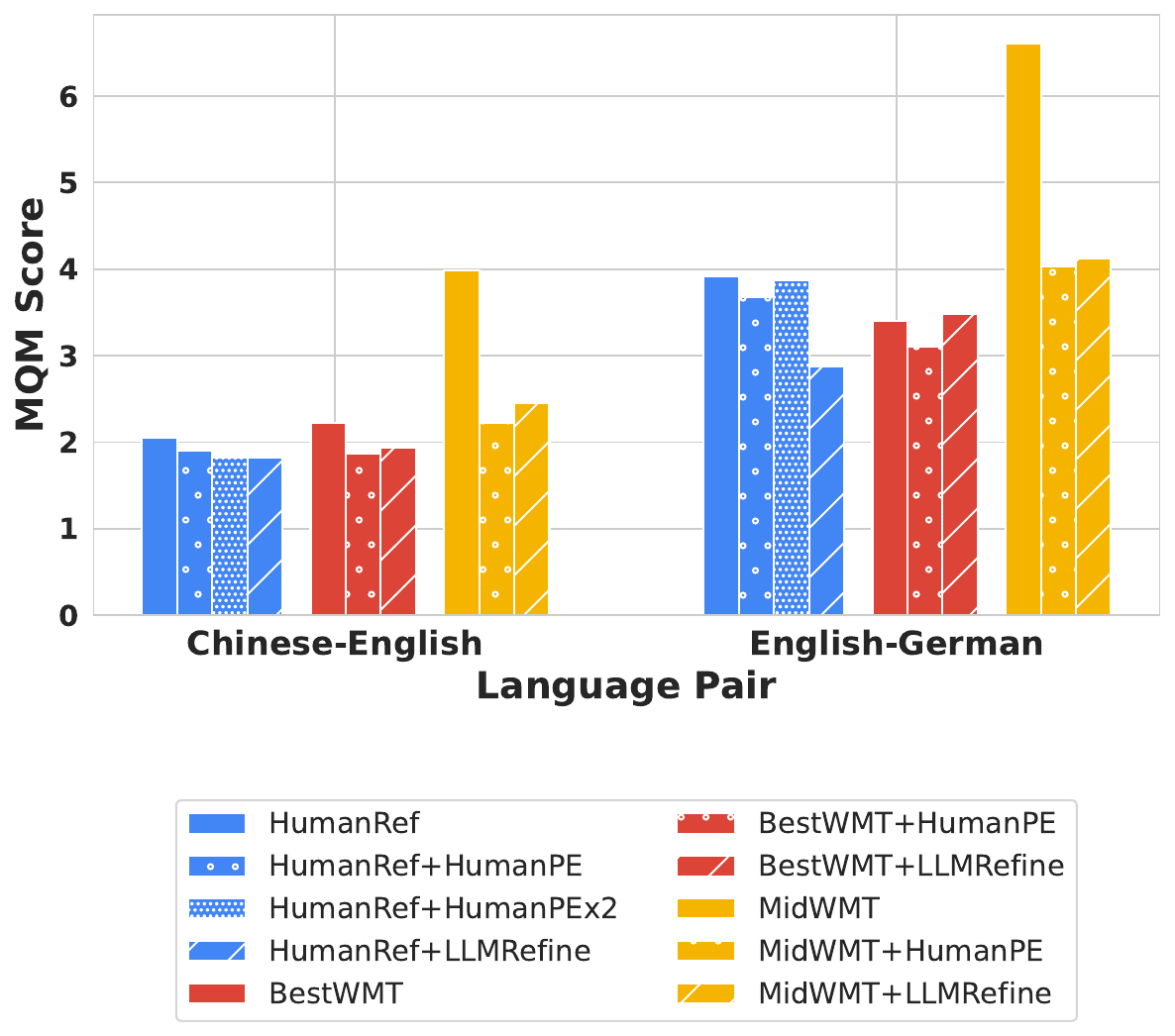}
  \caption{MQM Scores for different translation systems across two language pairs: Chinese-English and English-German. Bars represents the average MQM scores for each translation system. The systems are grouped and colored by initial translation and further categorized by post-editing method with different fill patterns. Lower MQM scores indicate better quality.}
  \label{fig:mqm_analysis}
\end{figure}


\begin{table*}[!]
    \centering
    \begin{adjustbox}{width=0.9\textwidth}
    \begin{tabular}{L{2cm}C{1cm}C{1.5cm}C{1.5cm}C{1.5cm}C{1.5cm}C{1.5cm}C{1.5cm}}
    \toprule
        \multicolumn{1}{l}{\bf Source} & \multicolumn{1}{c}{\bf Init. Translation} & \multicolumn{2}{c}{\bf + HumanPE} & \multicolumn{2}{c}{\bf + HumanPEx2} & \multicolumn{2}{c}{\bf + LLMRefine }\\
        \multicolumn{1}{l}{} & & \multicolumn{1}{c}{\bf Score $\downarrow$ } & \multicolumn{1}{c}{\bf $\Delta$} & \multicolumn{1}{c}{\bf Score $\downarrow$} & \multicolumn{1}{c}{\bf $\Delta$} & \multicolumn{1}{c}{\bf Score $\downarrow$} & \multicolumn{1}{c}{\bf $\Delta$}\\
        \midrule
        \textsc{En-De}  \\
        \midrule
        \textsc{HumanRef} & 3.79 & 3.53 & -0.26 & 3.70 & -0.09 & \bf 2.76 & \bf -1.03\\
        \textsc{BestWMT} & 3.30 & \bf 2.98 & \bf -0.32 & - & - & 3.36 & +0.06\\
        \textsc{MidWMT} & 6.45 & \bf 3.83 & \bf -2.62 & - & - &  4.02 & -2.43\\
        \midrule
        \textsc{Zh-En}  \\
        \midrule
        \textsc{HumanRef} & 2.05 & 1.91 & -0.14 & \bf 1.82 & \bf -0.23 & \bf 1.82  & \bf -0.23\\
        \textsc{BestWMT} & 2.22 & \bf 1.87 & \bf -0.35 & - & - & 1.94  & -0.28\\
        \textsc{MidWMT} & 3.98 & \bf 2.23 & \bf -1.75 & - & - & 2.45 & -1.53\\
        \bottomrule
    \end{tabular}
    \end{adjustbox}
    \caption{MQM human evaluation comparison of each post-edit approach on different initial translations. Lower MQM scores indicates better quality.}
    \label{tab:post-edit-compare}
\end{table*}

Table~\ref{tab:post-edit-compare} and Figure~\ref{fig:mqm_analysis} present a detailed analysis that highlights the quality benefits of human-machine collaboration.

Figure~\ref{fig:mqm_analysis} shows the gains in quality that can be provided by our post-edit approaches. The gains are more pronounced when starting with a lower-quality translation (\textsc{MidMT}), but even high-quality translations (\textsc{HumanRef}, \textsc{BestWMT}) can be improved. The quality differences between initial translations are greatly reduced after post-editing, but not eliminated.

Table~\ref{tab:post-edit-compare} provides a finer-grain analysis of the effect of each post-edit approach on different initial translations. In both language pairs, an LLM-based method provides the most benefit when starting with human translation, while human post-editing provides the most benefit for machine translation. Recall that in Table~\ref{tab:post-edit-cross-blue-compare} we showed that model-based methods and humans make more changes to translations from the other group; here we see that these changes are also net-positive. This indicates that \textbf{human-machine collaboration is an effective way to achieve high-quality translations}.

\section{Error Analysis}
\label{sec:error_analysis}
Here we present more detailed error analysis of both initial translation and post-edit stages to investigate where the quality improvements originate. Using English-German as an example, we first present analysis of the initial translations in Section~\ref{sec:error_analysis_init_translation} to understand the initial error distributions. Then, we further investigate the error-correction dynamics during post-editing in Section~\ref{sec:error_analysis_post_edit} to understand why human-machine collaboration stands out from other approaches. 

\subsection{Error distribution from Initial Translation}
\label{sec:error_analysis_init_translation}


\begin{table}[t]
    \centering
    \begin{adjustbox}{width=\columnwidth}
    \begin{tabular}{crrrr}
    \toprule
    \textbf{Error Type} & \textbf{OrigHumanRef}  & \textbf{HumanRef} & \textbf{BestWMT} & \textbf{MidWMT}\\
    \midrule
    \textsc{\bf No-error} & 200 & 175 & 144 & 116 \\
    \midrule
    \textsc{\bf Major} & 186 & 263 & 211 & 470 \\
    \midrule
    Fluency & 27 (15\%) & 29 (11\%) & 25 (12\%) & 88 (19\%) \\
    Accuracy & 114 (61\%) & 149 (57\%) & 108 (51\%) & 275 (59\%) \\
    Style & 22 (12\%) & 54 (21\%) & 45 (21\%) & 60 (13\%) \\
    \midrule
    \textsc{\bf Minor} & 659 & 745 & 891 & 1084 \\
    \midrule
    Fluency & 257 (39\%) & 242 (32\%) & 439 (49\%) & 538 (50\%) \\
    Accuracy & 181 (27\%) & 211 (28\%) & 186 (21\%) & 226 (21\%) \\
    Style & 141 (21\%) & 190 (26\%) & 178 (20\%) & 213 (20\%) \\
    \bottomrule
    \end{tabular}
    \end{adjustbox}
    \caption{Error type and severity distributions of English-German MQM human evaluation results.}
    \label{tab:ende-init-translation}
\end{table}

We present the error type and severity distribution of the English-German initial translations in Table~\ref{tab:ende-init-translation} and that of Chinese-English in Appendix~\ref{sec:error_analysis_init_translation_appendix}. It shows that accuracy-related errors are the primary source of major errors and that the distribution of minor errors is more evenly spread across different categories. Importantly, these error distributions are similar for both human and machine-based initial translations. This consistency provides a solid foundation for comparing different post-editing techniques in our downstream analysis. 

\subsection{Error Correction from Post-Editing}
\label{sec:error_analysis_post_edit}

To understand the error-correction dynamics of different post-editing methods and find the origin of the improvements of human-machine collaboration, we explored three key questions:
\begin{itemize}
    \item  Do different post-editing methods agree on which segments to modify?
    \item  How do different post-editing methods affect the total number of major and minor errors?
    \item  How do different post-editing methods affect the total number of high- and low-quality segments?
\end{itemize}

\begin{figure}[t]
  \includegraphics[width=0.99\columnwidth]{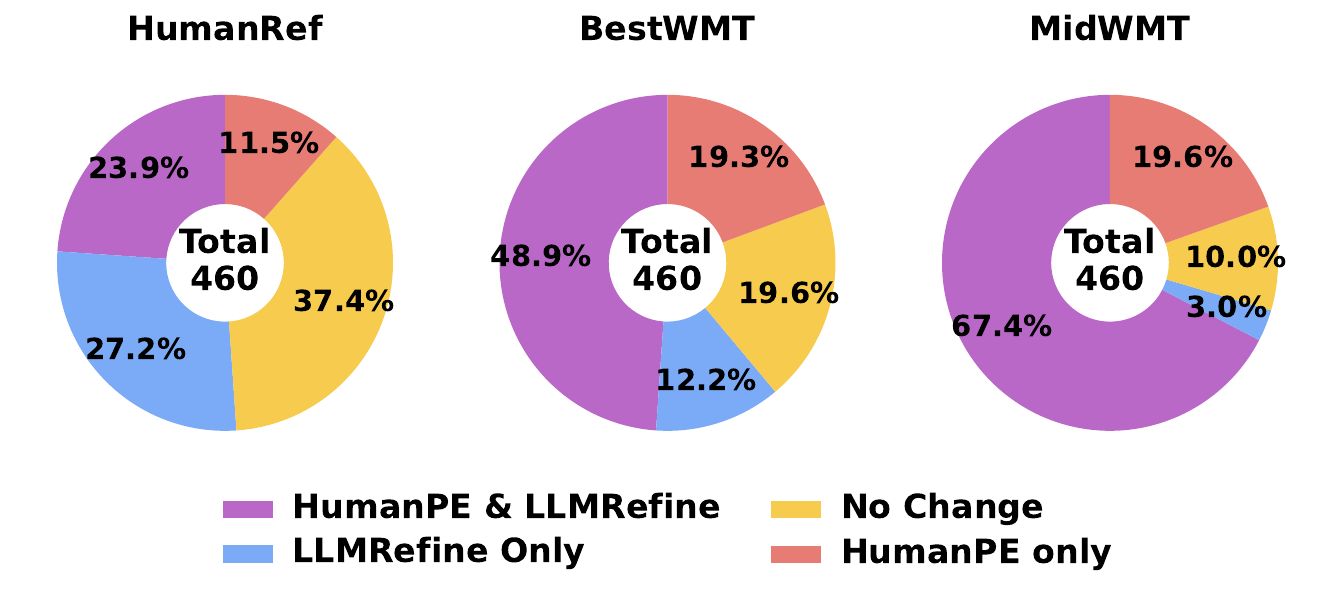}
  \caption{Agreement between HumanPE and LLMRefine in identifying segments requiring post-edit on English-German data. Each pie chart\protect\footnotemark represents a different initial translation source.}
  \label{fig:ende_post_edit_agreement}
\end{figure}

We first examine how often human post-editors (\textsc{HumanPE}) and machine post-editing methods (\textsc{LLMRefine})  agree on which segments need correction. Figure~\ref{fig:ende_post_edit_agreement} shows that both methods identify more segments for editing in lower-quality initial translations as evidenced by the shrinking "No Change" (yellow) section. Notably, agreement between \textsc{HumanPE} and \textsc{LLMRefine} increased from 23.9{\%} for high-quality \textsc{HumanRef} translations to 67.4{\%} for lower-quality MidWMT translations as observed by the expanded "HumanPE \& LLMRefine" (purple) section. This suggests more consensus on obvious errors in lower-quality texts while greater divergence for higher-quality translations in editing approaches. A detailed numerical breakdown is shown in Table~\ref{tab:ende-post-edit-agree} in Appendix~\ref{sec:error_analysis_post_editing}. 

\footnotetext{Tabular statistics are provided in Tables~\ref{tab:ende-post-edit-agree} and ~\ref{tab:zhen-post-edit-agree} in Appendix~\ref{sec:error_analysis_post_editing}.}

Another interesting observation from Figure~\ref{fig:ende_post_edit_agreement} is that \textsc{HumanPE} (purple + red) identifies a larger proportion of segments needing correction in \textsc{BestWMT} (48.9\%  + 19.3\% = 68.2\%) compared to \textsc{HumanRef} (23.9\%  + 11.5\% = 35.4\%), despite the superior quality of \textsc{BestWMT} over \textsc{HumanRef} as shown in Table~\ref{tab:ende_mqm}. This suggests that human post-editors might overlook certain errors in human translations due to their familiar patterns. Conversely, the unfamiliar patterns in machine-generated text may make errors more salient to human editors. This interpretation is consistent with the pattern depicted in Figure~\ref{fig:cross_bleu}, where human post-editors make fewer changes to human translations than to machine translations. This observation provides one plausible explanation for the necessity of human-machine collaboration in achieving high-quality translations. Detailed statistics for English-German are shown in Table~\ref{tab:ende-post-edit-agree} and similar trends are also observed in Chinese-English in Table~\ref{tab:zhen-post-edit-agree} in Appendix~\ref{sec:error_analysis_post_editing}.

\begin{figure}[t]
  \includegraphics[width=0.95\columnwidth]{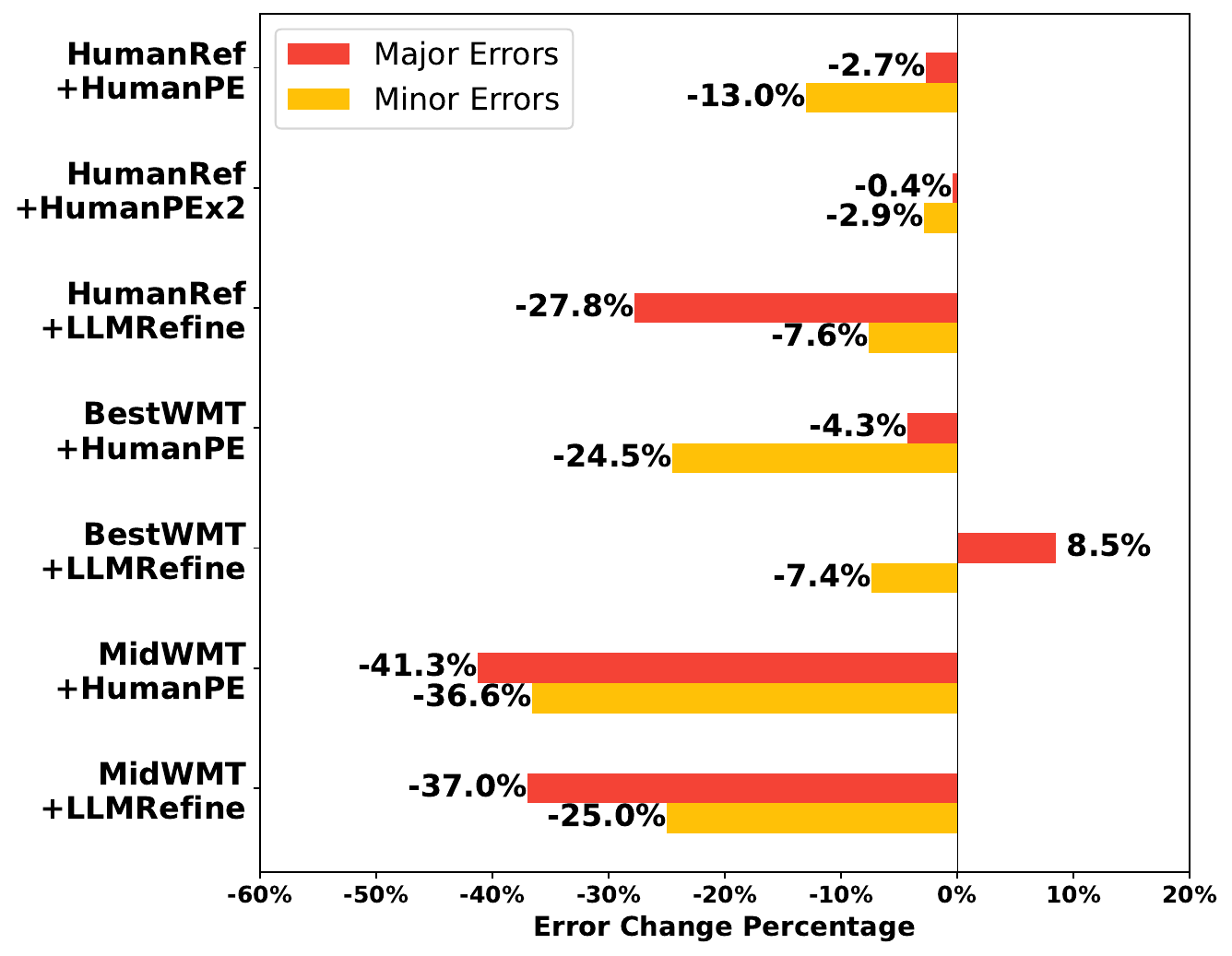}
  \caption{Error changes percentages by different post-editing approaches on English-German data. The percentages present the changes in error counts for each post-editing method compared to its initial translation. A negative indicates a decrease in errors, while positive value indicates an increase in the error type.}
  \label{fig:ende_post_edit_severity}
\end{figure}

We wish to investigate how many errors are corrected during post-editing, but because it is difficult to automatically determine whether an individual error was corrected, we instead examine how the total number of errors changes after post-editing, also considering severity. Figure~\ref{fig:ende_post_edit_severity} shows that both human and machine post-editing reduce overall error counts across different initial translation qualities. \textsc{LLMRefine} outperforms \textsc{HumanPE} on ~\textsc{HumanRef} initial translation in reducing major errors (-27.8\% vs. -2.7\%), while \textsc{HumanPE} is superior for ~\textsc{BestWMT}, decreasing major-error segments compared to the increase for \textsc{LLMRefine} (-4.3\% vs. +8.5\%). On \textsc{MidWMT}, both methods show substantial improvements, with \textsc{HumanPE} moderately ahead of \textsc{LLMRefine} (-41.3\% vs. -37.0\% decrease in major-error segments). These findings highlight the complementary strengths of human and machine post-editing methods, indicating that a hybrid method is likely the most effective strategy for reducing errors, regardless of the initial translation's origin. Similar trends are also observed in Chinese-English in Figure~\ref{fig:zhen_post_edit_severity} in Appendix~\ref{sec:error_analysis_post_editing}.

\begin{figure}[t]
    \centering
    \begin{subfigure}{0.9\columnwidth}
        \centering
        \includegraphics[width=0.95\linewidth]{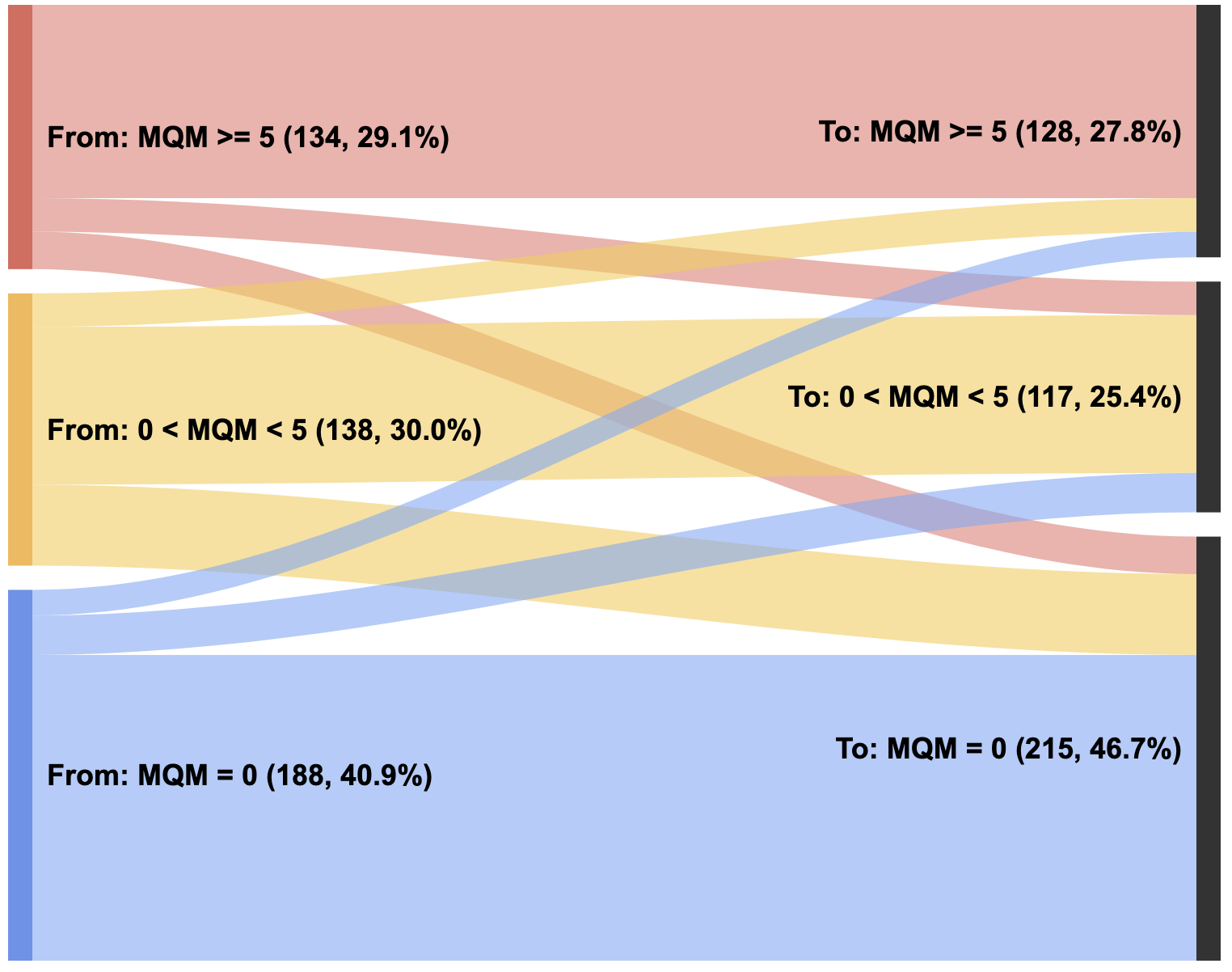}
        \caption{From \textsc{HumanRef} to \textsc{HumanRef+HumanPE}}
        \label{fig:ende_sankey_newref_e1}
    \end{subfigure}
    \begin{subfigure}{0.9\columnwidth}
        \centering
        \includegraphics[width=0.95\linewidth]{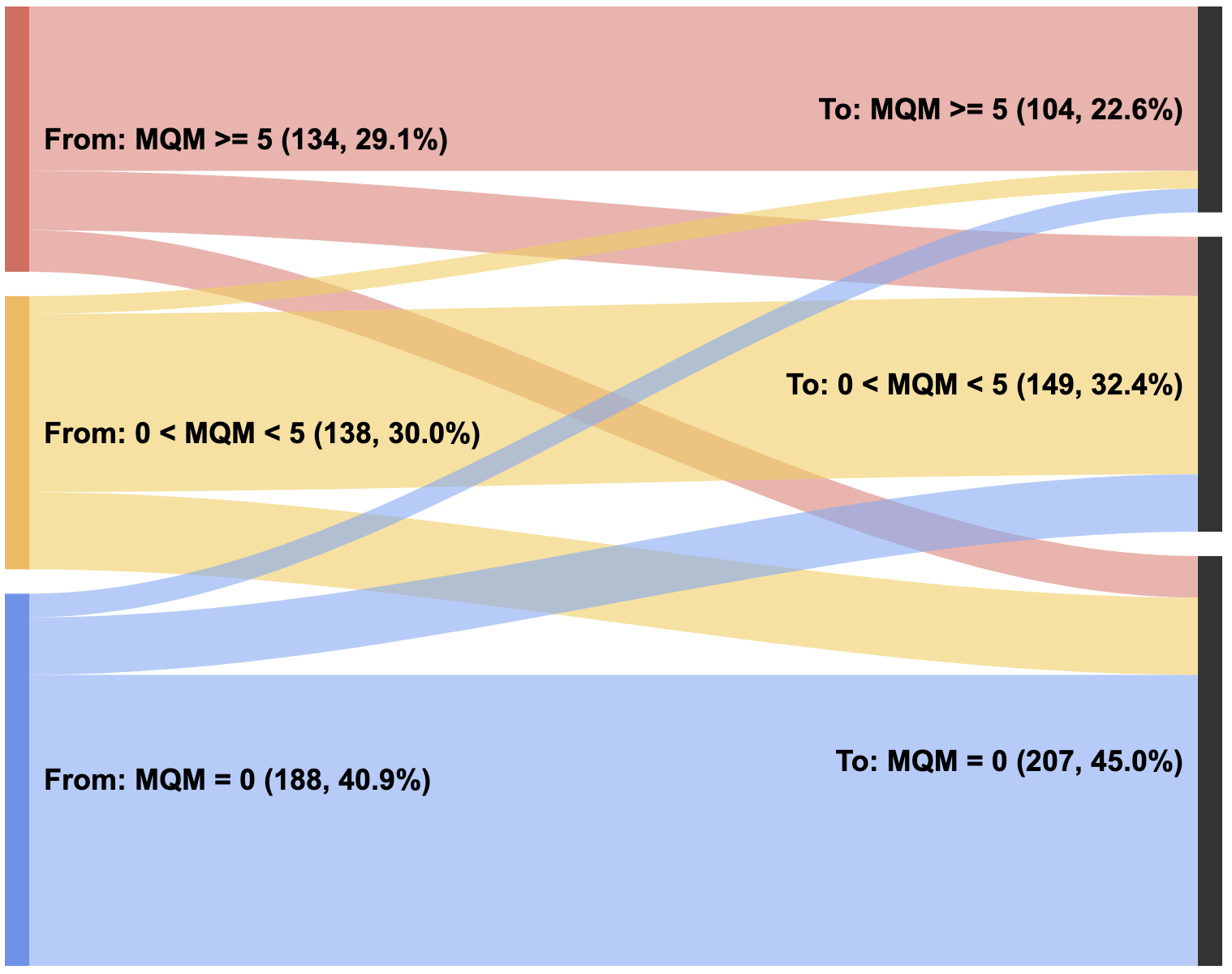}
        \caption{From \textsc{HumanRef} to \textsc{HumanRef+LLMRefine}}
        \label{fig:ende_sankey_newref_llmrefine}
    \end{subfigure}
    \caption{Segment-level quality shift through \textsc{HumanPE} and \textsc{LLMRefine} from English-German \textsc{HumanRef}. Each segment is categorized into one of three groups based on its MQM score: 1) high-scoring segments with MQM >= 5; 2) low-scoring segments with 0 < MQM < 5; 3) error-free segments with MQM=0. Higher MQM scores indicate more numerous/severe errors and accordingly lower translation quality.}
    \label{fig:sankey_error_flow_ende}
\end{figure}
\begin{figure}[t]
    \centering
    \begin{subfigure}{0.9\columnwidth}
        \centering
        \includegraphics[width=0.95\linewidth]{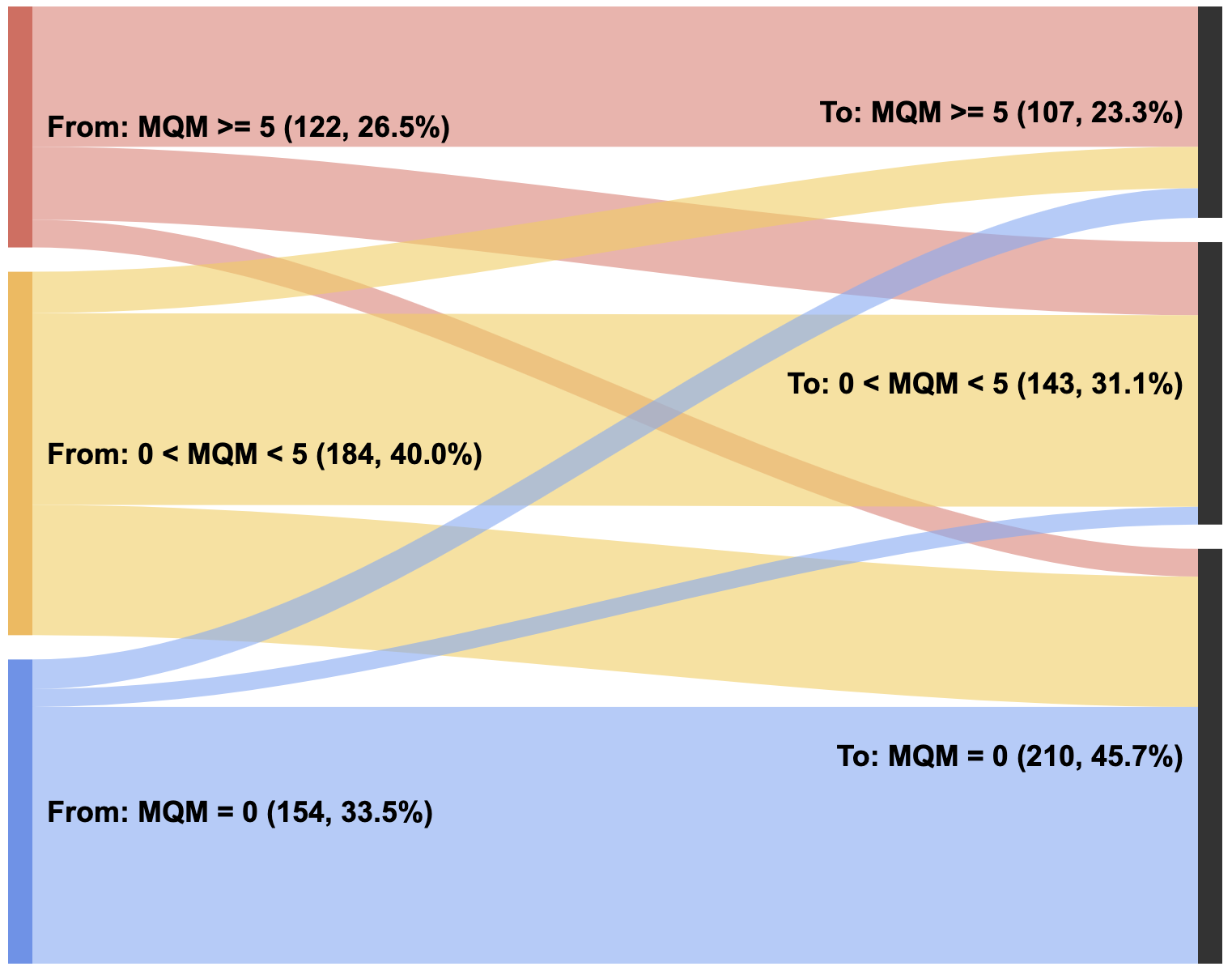}
        \caption{From \textsc{BestWMT} to \textsc{BestWMT+HumanPE}}
        \label{fig:ende_sankey_bestwmt_pe}
    \end{subfigure}
    \begin{subfigure}{0.9\columnwidth}
        \centering
        \includegraphics[width=0.95\linewidth]{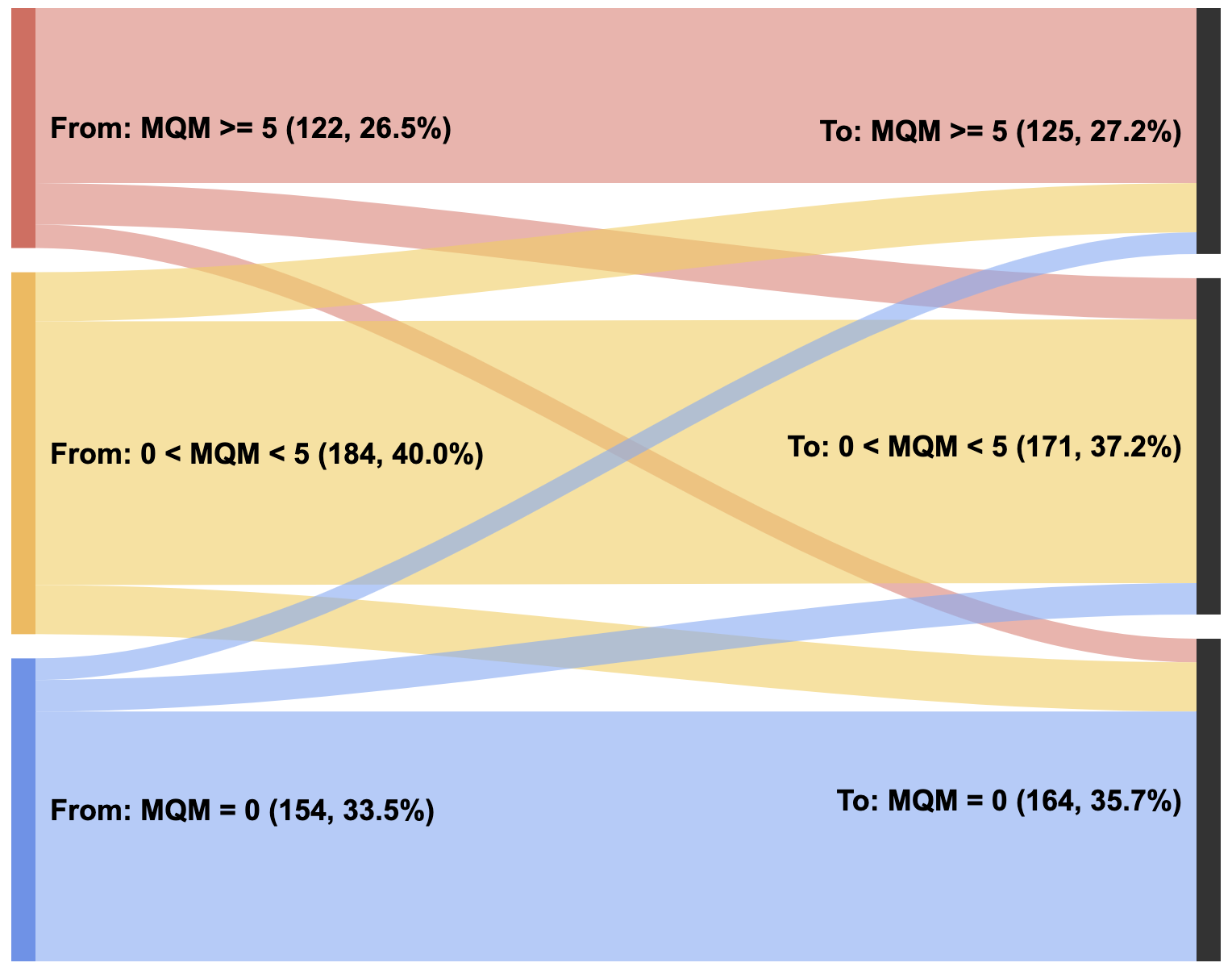}
        \caption{From \textsc{BestWMT} to \textsc{BestWMT+LLMRefine}}
        \label{fig:ende_sankey_bestwmt_llmrefine}
    \end{subfigure}
    \caption{Segment-level  quality  shift  through \textsc{HumanPE} and \textsc{LLMRefine} from English-German \textsc{BestWMT}.}
    \label{fig:sankey_error_flow_ende_bestwmt}
\end{figure}

To understand the error correction dynamics for each segment, we analyzed how MQM scores change before and after post-editing. Ideally, post-editing would fix existing errors while minimizing the introduction of new ones. However, as Figure~\ref{fig:sankey_error_flow_ende} illustrates, post-editing is not guaranteed to improve every segment: while some segments are improved, others are worsened.

Figure~\ref{fig:sankey_error_flow_ende} compares \textsc{HumanPE} and \textsc{LLMRefine} on \textsc{HumanRef} initial translations for English-German data. Both methods reduce the number of high-scoring (low-quality) segments (MQM >= 5). Notably, \textsc{LLMRefine} outperforms \textsc{HumanPE} by showing fewer quality-degrading corrections and more quality-improving ones. \textsc{LLMRefine} minimizes low-to-high-scoring transitions with a narrower flow from low-scoring segments (0 < MQM < 5) to high-scoring segments compared to \textsc{HumanPE}. Moreover, \textsc{LLMRefine} achieves a significant reduction in high-scoring segments by 6.5\% (from 29.1\% to 22.6\%) compared to \textsc{HumanPE}'s 1.3\% (from 29.1\% to 27.8\%), suggesting that it is more effective at achieving post-editing gains while preserving originally good translations. A similar trend is observed for Chinese-English with \textsc{HumanRef} initial translation in Figure~\ref{fig:sankey_error_flow_zhen} in Appendix~\ref{sec:error_analysis_post_editing}.

To demonstrate that the quality improvement is not solely due to the capabilities of \textsc{LLMRefine}, we conducted further experiments as shown in Figure~\ref{fig:sankey_error_flow_ende_bestwmt}. This figure compares \textsc{HumanPE} and \textsc{LLMRefine} with \textsc{BestWMT} initial translations. Interestingly, the results are reversed: \textsc{HumanPE} outperforms \textsc{LLMRefine} in this scenario, showing fewer quality-degrading corrections and more quality-improving ones. \textsc{HumanPE} demonstrates a significantly wider flow from high-scoring segments to low-scoring ones. It achieves a notable reduction in high-scoring segments by 3.2\% (from 26.5\% to 23.3\%), while \textsc{LLMRefine} sees an increase of 0.7\% (from 26.5\% to 27.2\%). Furthermore, \textsc{HumanPE} significantly increases the No-Error segments (MQM=0) by 12.2\% (from 33.5\% to 45.7\%) compared to \textsc{LLMRefine}'s 2.2\% increase.

The distinct performance differences of \textsc{LLMRefine} and \textsc{HumanPE} in these two sets of experiments highlight that \textbf{the quality improvement stems primarily from the complementary strengths of human and machine collaboration}, rather than the superior capability of either \textsc{LLMRefine} or \textsc{HumanPE} alone. This underscores the importance of leveraging both human and machine strengths in achieving optimal translation quality.





\section{Costs Analysis}
\label{sec:costs_analysis}

\begin{table}[t]
    \centering
    \begin{adjustbox}{width=0.98\columnwidth}
    \begin{tabular}{l|cc|cc|c}
        \toprule
        \multicolumn{1}{l}{\bf Systems} & \multicolumn{2}{c}{\bf Quality Rank} & \multicolumn{1}{c}{\bf Costs} \\
        \multicolumn{1}{c}{\bf }& \multicolumn{1}{c}{\bf EnDe} & \multicolumn{1}{c}{\bf ZhEn} & \multicolumn{1}{c}{\bf }\\
        \midrule
        \textsc{HumanRef} & 3 & 2 & 1X\\
        \textsc{HumanRef+HumanPE} & 2 & 1 & 1.6X\\
        \textsc{HumanRef+HumanPEx2} & 2 & 1 & 2.2X\\
        \textsc{HumanRef+LLMRefine} & \bf 1 & \bf 1 & 1X\\
        \textsc{BestWMT+HumanPE} &  \bf 1 &  \bf 1 & \bf 0.6X\\
        \bottomrule
    \end{tabular}
    \end{adjustbox}
    \caption{Quality rank and costs comparison of different data collection systems. 1st rank indicates the translation quality belongs to the highest quality significance cluster in Table~\ref{tab:ende_mqm} and ~\ref{tab:zhen_mqm}.}
    \label{tab:costs_analysis}
\end{table}

So far we have focused on comparing quality between various translation data collection approaches. However, practical considerations make it important to consider the trade-off between quality and costs. Table~\ref{tab:costs_analysis} analyzes relative human annotation costs between various approaches, along with the rank of the significance cluster that each method appeared in. The exact costs for the human annotation conducted in this study are confidential (although all annotators were paid fair market wages), so we instead use relative costs, based on the industry standard that post-editing text of a given length takes less time (and accordingly costs less) than producing a translation of that length. We specifically assume that human post-editing costs around 60{\%} of what human translation does. According to existing literature ~\citep{Plitt2010APT,zouhar-etal-2021-neural,human-pe-efficacy-2013} and internal statistics, we believe it's a fair assumption, although the exact costs can vary upon different vendors, languages, task size, etc.

With this framework, the best combination of quality and cost appears to be human post-editing of high-quality MT (\textsc{BestWMT+HumanPE}), attaining quality in the top significance cluster in both language pairs with only 60{\%} of the human annotation cost of collecting an initial human translation.
Meanwhile, we see that one or two rounds of human post-editing of an initial human translation increases costs without a meaningful gain in quality, while just applying an LLM post-editor (\textsc{HumanRef+LLMRefine}) brings quality to the top significant cluster with no additional human annotation cost, making it a viable option when human translations are already collected. It's worth noting that LLM inference costs are negligible (on the order of dollars per million tokens) compared to human annotation costs, further enhancing the cost-effectiveness of LLM-based approaches. This indicates that \textbf{human-machine collaboration can be a faster, more cost-efficient alternative} to traditional collection of translations from humans, optimizing both quality and resource allocation by leveraging the strengths of both humans and machines.



\section{Related Work}
\label{sec:related_work}
There have been a few studies investigating methods of acquiring high-quality translations. Recently, \citet{zouhar2024evaluating} proposed collecting high-quality translations by building consensus between multiple translators. \citet{zouhar2024quality} proposed collecting multiple translations from different tiers of human translators with careful budget calculations to optimize cost-efficiency.

\paragraph{Human Post-Edits}
Computer-aided translation tools are now widely used by professional translators for interactive translation and post-editing~\citep{alabau2014casmacat,federico2014matecat,green-etal-2014-human,denkowski2015machine,sin2014development, kenny2012electronic}.
\citet{carl2011process} have shown that human translators work faster and make fewer mistakes when editing machine translations than when translating from scratch. \citet{toral2018post} supports this, demonstrating even greater improvements with neural machine translation compared to phrase-based systems. \citet{zouhar-etal-2021-neural} investigates the relationship between machine translation quality and post-editing efforts and found no straightfoward relationship.
On the other hand, \citet{popovic-etal-2016-potential} suggested that post-edits should be used carefully for MT evaluation due to the bias of each post-edit towards its MT system.
Further, \citet{toral-2019-post} showed that human post-edits are simpler and more normalised in language than human translations from scratch.
 

\paragraph{Automatic Refinement}
\citet{lin-etal-2022-automatic} showed how the
errors that humans make differ from those made by
MT systems. They constructed a Translation Error Correction (TEC) corpus with professional translators and showed that models trained on it outperform Automatic Post-Editing (APE) models \citep{knight1994automated} that are trained to correct MT output.
Since the emergence of LLMs, new refinement approaches based on detailed MQM annotations have appeared \citep{xu-etal-2023-instructscore, fernandes-etal-2023-devil}. \citet{xu2024llmrefine} showed that these refinement method can be used to improve the quality of human translations.

Meanwhile, machines have been extensively evaluated and utilized as an alternative to human annotators for data collection \citep{zouhar-etal-2021-neural,yan2024gpt4vshumantranslators}.


In contrast to the above methods, we investigate the interaction between humans and machines in the initial translation and post-editing stages, including detailed analysis of the resulting changes in quality while also considering cost-efficiency.

\section{Conclusion}
\label{sec:conclusion}

We investigate various approaches for gathering translation data, including human-only, machine-only, and hybrid approaches. Our results demonstrate that human-machine collaboration can consistently generate high-quality translations at a lower cost than human-only methods. Through detailed error analysis, we uncovered the nuances of error correction dynamics and highlighted the advantages of human-machine collaborative methods. Our cost analysis also demonstrates the cost-efficiency of human-machine collaboration methods. Finally, we release to the public a dataset of roughly 18,000 translation segments of varying quality from different collection methods along with human ratings, to facilitate further research in this area.

\newpage

\section*{Limitations}
\label{sec:limitation}
This study focuses on two language pairs, English-German and Chinese-English. They are chosen due to the extensive study in the WMT23 metrics shared task \citep{freitag-etal-2023-results} and the availability of data from various translation systems from the WMT23 general shared task \citep{kocmi-etal-2023-findings}. While our analysis provides support for the findings presented in this work and we offer a plausible explanation for the observed results, it is important to acknowledge certain variables are not accounted for in this work, including using translators or post-editors with varying quality levels, different systems for translation and post-editing, utilizing sentence or paragraph datasets from other domains, and higher or lower resource language pairs beyond the two investigated here. Therefore, we cannot guarantee the observed trends will generalize to different datasets. 

We want to especially highlight the need for further exploration of the quality variance observed among human translators, such as \textsc{OrigHumanRef} and \textsc{HumanRef} in the English-German translation task. The current study's limited annotation budget and timeline restricted the depth of this investigation. Future experiments aimed at examining the impact of post-editing on annotator agreement would be particularly interesting and valuable. 


\section*{Ethical Statement}
\label{sec:ethics_statement}
The source data used for translation and post-edits is accessible to the public. We're certain that the data annotated by human labors is free from risk or toxic content. We used an internal, proprietary tool to collect human translation, post-edits, and evaluation data. The annotators were compensated fairly and were not required to disclose any personal details during the annotation process. All the test data used in this study are publicly available and annotators were allowed to label sensitive information if necessary. The annotators are fully informed that the data they collected will be used for research purposes. 

\bibliography{anthology,paper}

\appendix

\section{Appendix}
\label{sec:appendix}

\subsection{Cross-BLUE scores}
\label{sec:cross_bleu_appendix}
\begin{figure}[!htb]
  \includegraphics[width=\columnwidth]{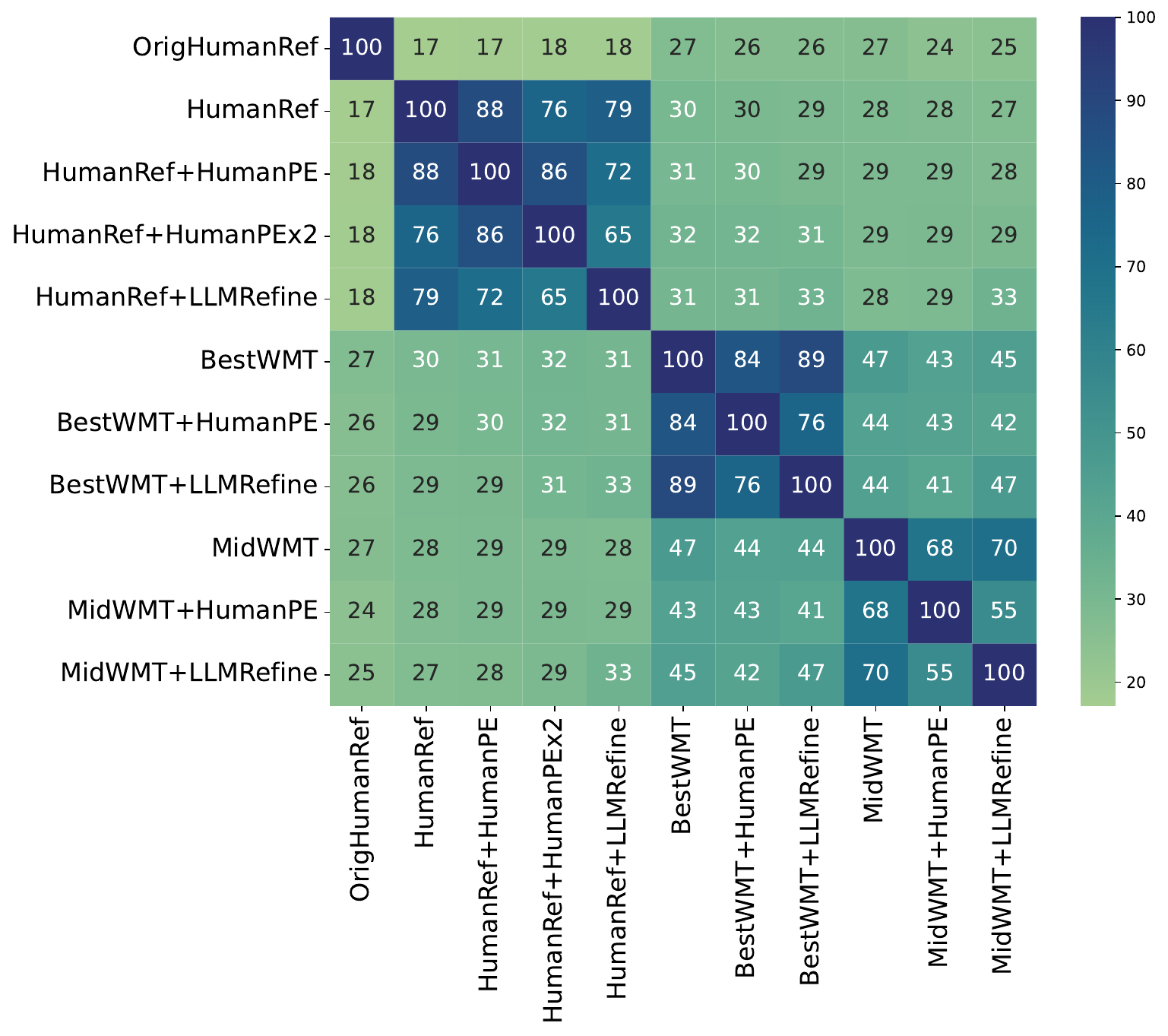}
  \caption{Cross-BLEU scores for different Chinese-English translation collection approaches}
  \label{fig:cross_bleu_zhen}
\end{figure}

Figure ~\ref{fig:cross_bleu_zhen} presents the cross-BLEU similarity matrix for Chinese-English translation systems.


\subsection{Error Distribution of Initial Translation}
\label{sec:error_analysis_init_translation_appendix}
Table ~\ref{table:zhen-init-translation} presents the error type and severity distributions of Chinese-English MQM human evaluation results.
\begin{table}[!htb]
    \centering
    \begin{adjustbox}{width=\columnwidth}
    \begin{tabular}{crrrr}
    \toprule
    \textbf{Error Type} & \textbf{OrigHumanRef}  & \textbf{HumanRef} & \textbf{BestWMT} & \textbf{MidWMT}\\
    \midrule
     \bf No-error & 225 & 453 & 490 & 304 \\
    \midrule
    \textsc{\bf Major} \\
    \midrule
    Total & 955 & 284 & 260 & 648 \\
    \midrule
    Fluency & 18 (2\%) & 15 (5\%) & 5 (2\%) & 23 (4\%) \\
    Accuracy & 851 (89\%) & 229 (81\%) & 221 (85\%) & 568 (88\%) \\
    Style & 31 (3\%) & 12 (4\%) & 5 (2\%) & 18 (3\%) \\
    \midrule
    \textsc{\bf Minor} \\
    \midrule
    Total & 1902 & 1303 & 1238 & 1704 \\
    \midrule
    Fluency & 625 (33\%) & 431 (33\%) & 364 (29\%) & 522 (31\%) \\
    Accuracy & 599 (31\%) & 412 (32\%) & 388 (31\%) & 447 (26\%) \\
    Style & 629 (33\%) & 402 (31\%) & 436 (35\%) & 677 (40\%) \\
    \bottomrule
    \end{tabular}
    \end{adjustbox}
    \caption{Error type and severity distributions of Chinese-English MQM human evaluation results.}
    \label{table:zhen-init-translation}
\end{table}

\subsection{Error Correction from Post-Editing}
\label{sec:error_analysis_post_editing}
\begin{figure}[htb!]
  \includegraphics[width=0.95\columnwidth]{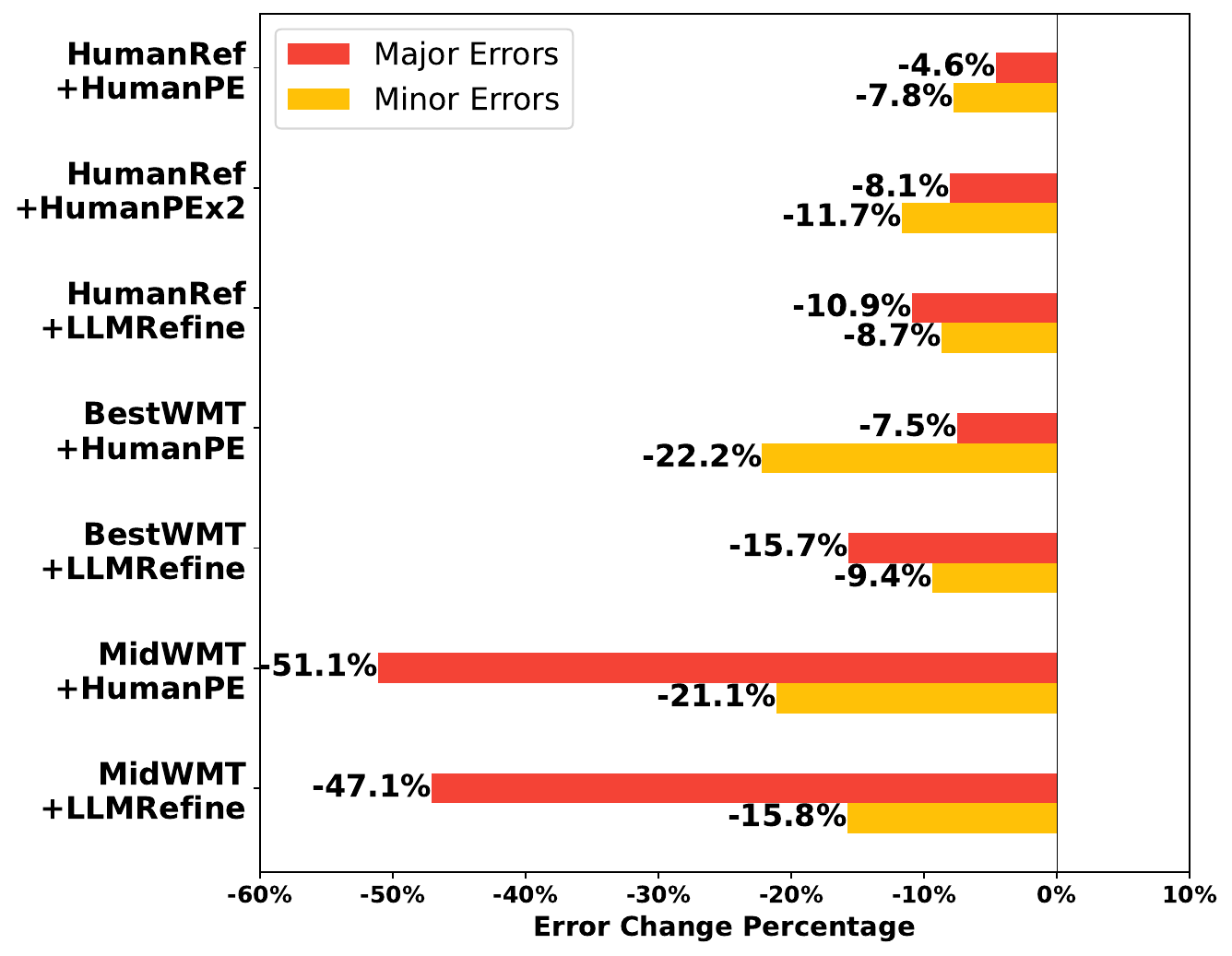}
  \caption{Error changes percentages by different post-editing approaches on Chinese-English data. The percentages present the changes in error counts for each post-editing method compared to its initial translation. A negative indicates a decrease in errors, while positive value indicates an increase in the error type.}
  \label{fig:zhen_post_edit_severity}
\end{figure}

\begin{figure}[!htb]
    \centering
    \begin{subfigure}{0.9\columnwidth}
        \centering
        \includegraphics[width=0.95\linewidth]{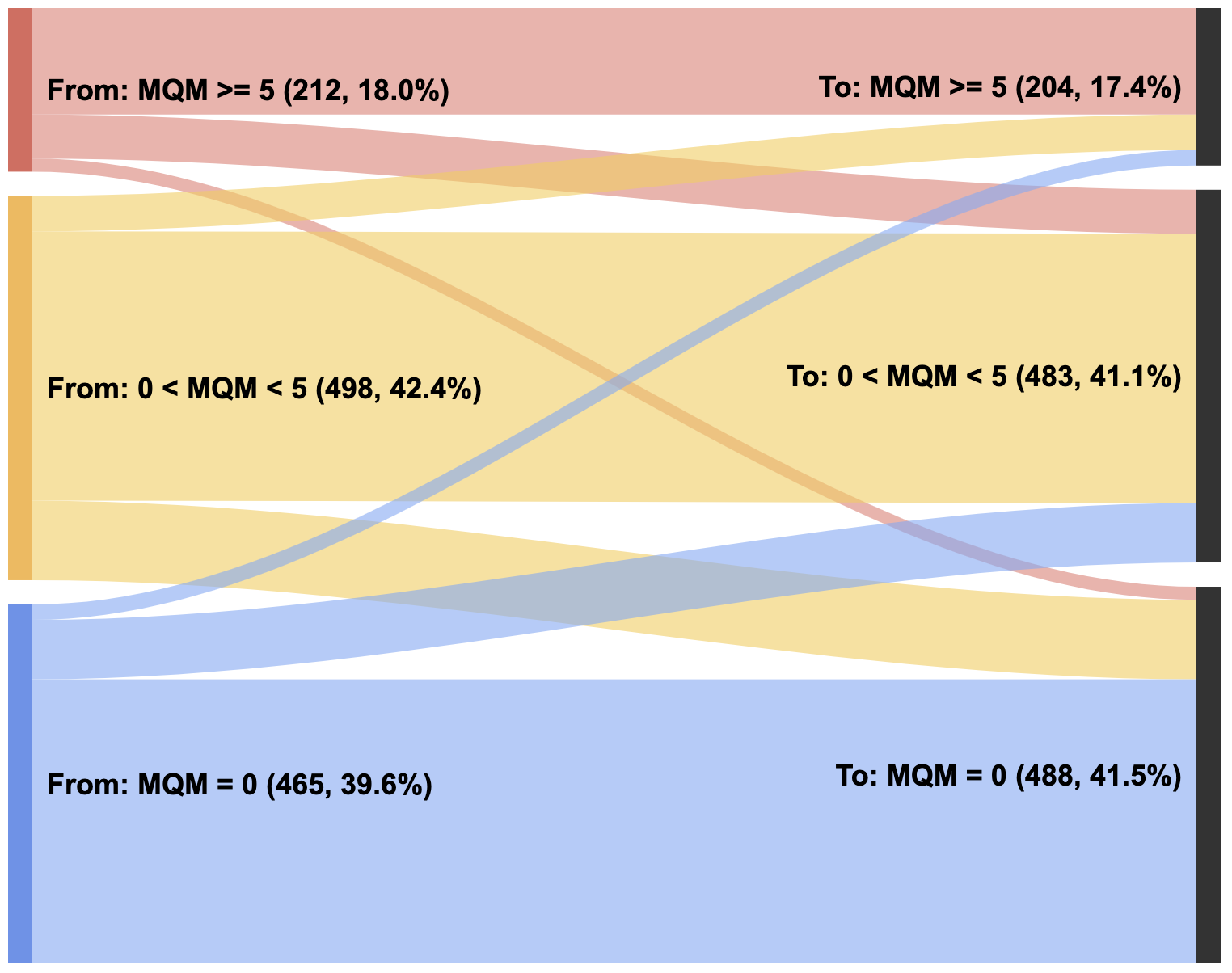}
        \caption{From \textsc{HumanRef} to \textsc{HumanRef+HumanPE}}
        \label{fig:zhen_sankey_newref_e1}
    \end{subfigure}
    \begin{subfigure}{0.9\columnwidth}
        \centering
        \includegraphics[width=0.95\linewidth]{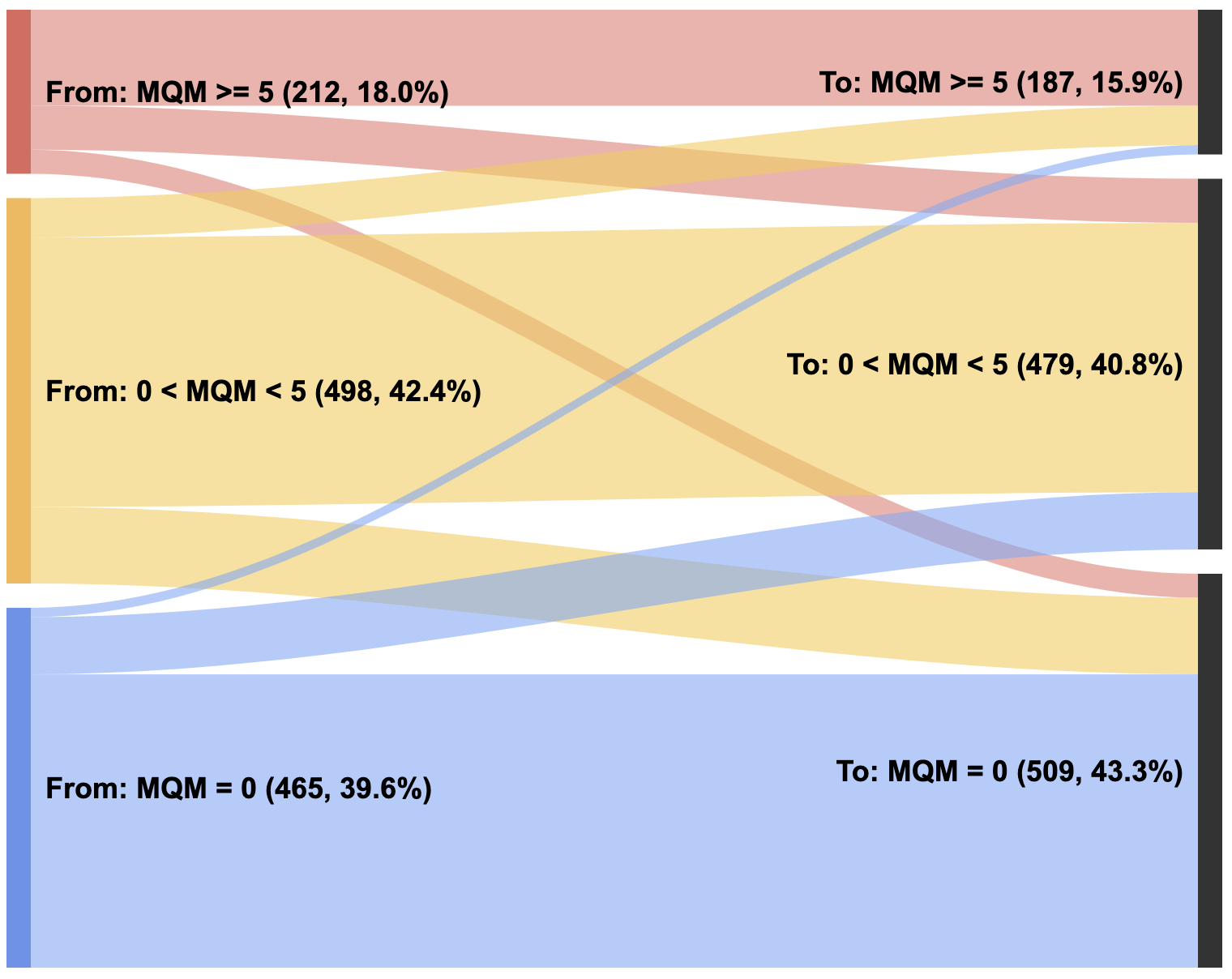}
        \caption{From \textsc{HumanRef} to \textsc{HumanRef+LLMRefine}}
        \label{fig:zhen_sankey_newref_llmrefine}
    \end{subfigure}
    \caption{Segment-level  quality  shift  through \textsc{HumanPE} and \textsc{LLMRefine} from Chinese-English \textsc{HumanRef}.}
    \label{fig:sankey_error_flow_zhen}
\end{figure}

\begin{figure}[!htb]
    \centering
    \begin{subfigure}{0.9\columnwidth}
        \centering
        \includegraphics[width=0.95\linewidth]{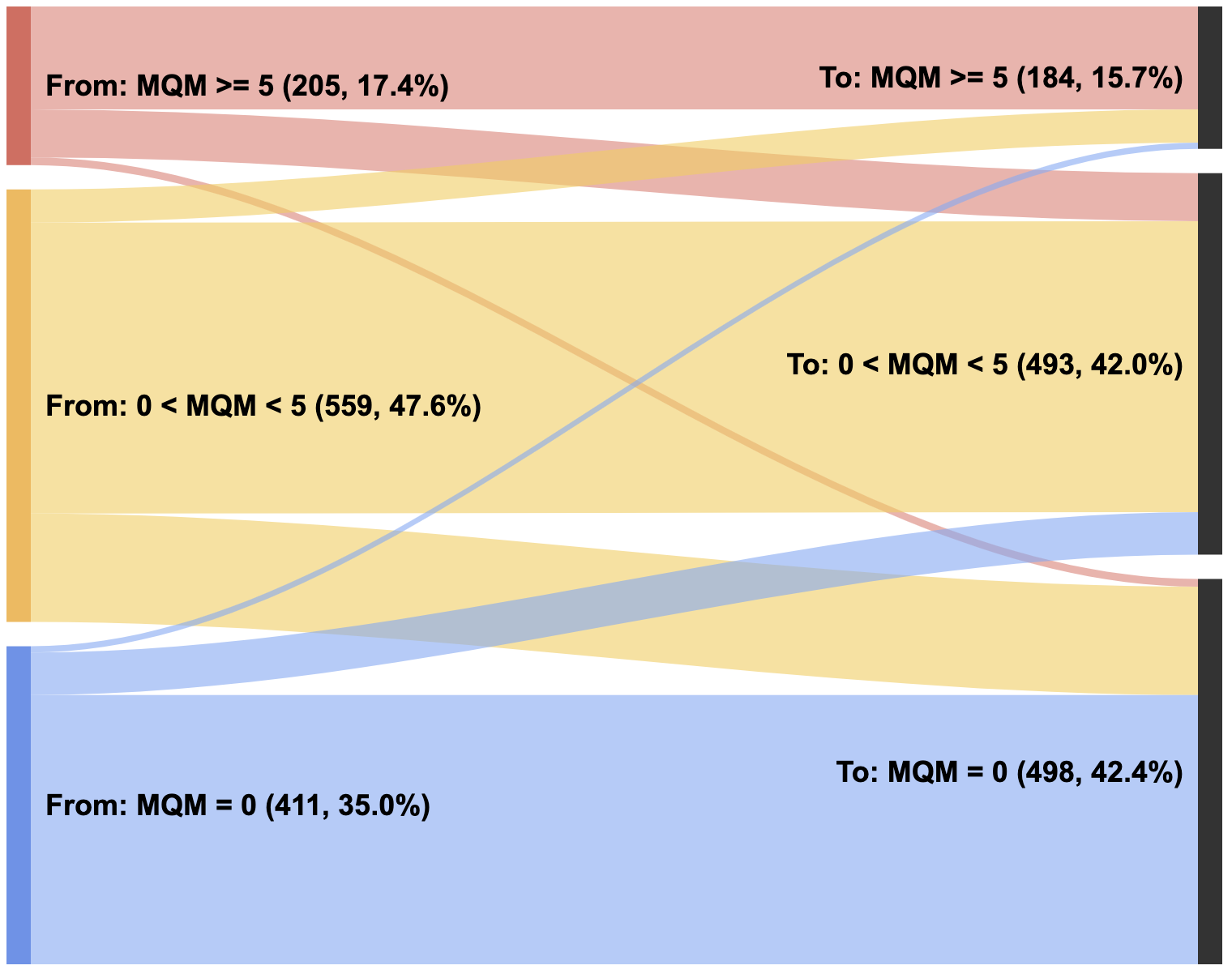}
        \caption{From \textsc{BestWMT} to \textsc{BestWMT+HumanPE}}
        \label{fig:zhen_sankey_bestwmt_pe}
    \end{subfigure}
    \begin{subfigure}{0.9\columnwidth}
        \centering
        \includegraphics[width=0.95\linewidth]{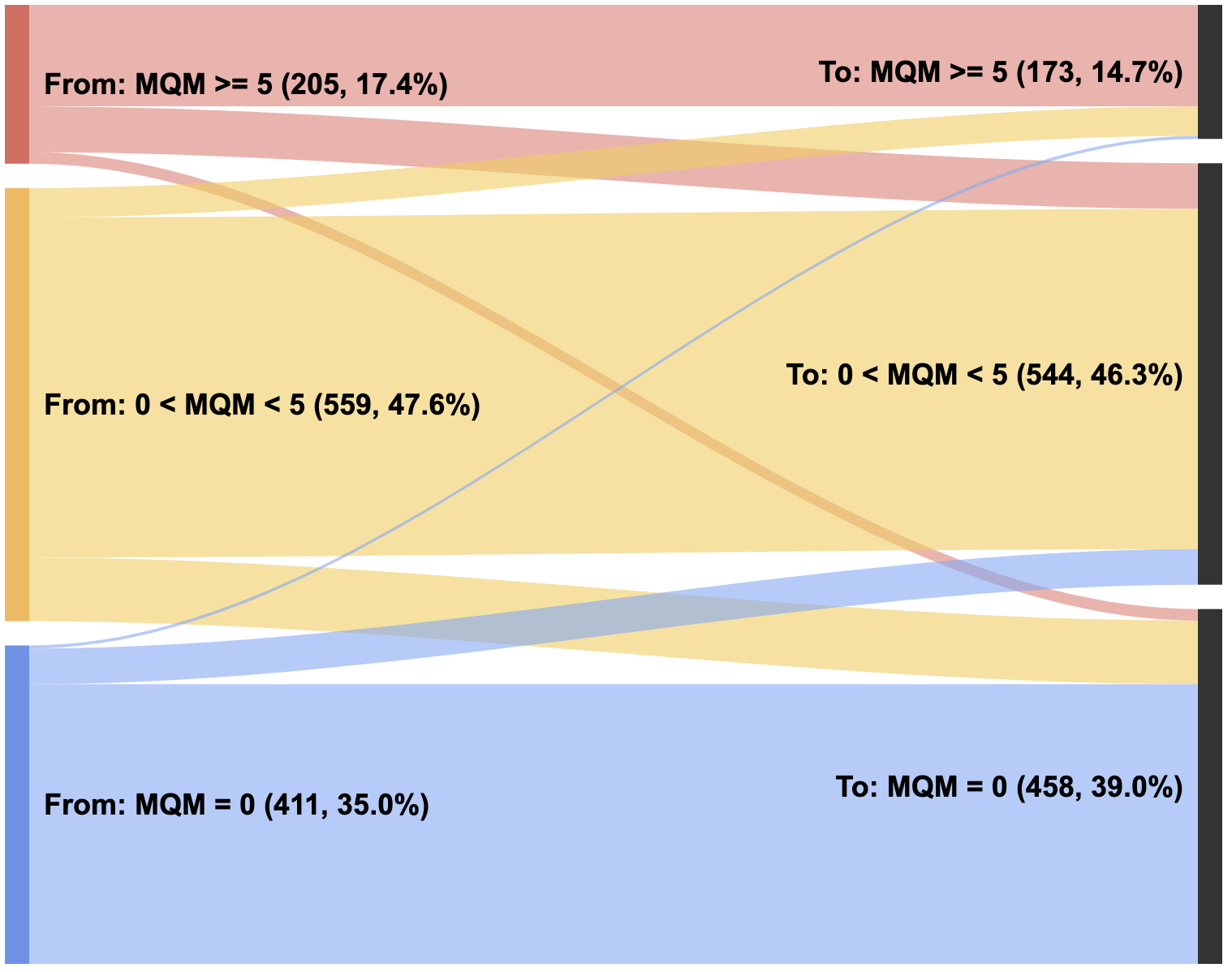}
        \caption{From \textsc{BestWMT} to \textsc{BestWMT+LLMRefine}}
        \label{fig:zhen_sankey_bestwmt_llmrefine}
    \end{subfigure}
    \caption{Segment-level  quality  shift  through \textsc{HumanPE} and \textsc{LLMRefine} from Chinese-English \textsc{BestWMT}.}
    \label{fig:sankey_error_flow_zhen_bestwmt}
\end{figure}

Figures~\ref{fig:zhen_post_edit_severity},~\ref{fig:sankey_error_flow_zhen}, and~\ref{fig:sankey_error_flow_zhen_bestwmt} present Chinese-English results comparable to the English-German results presented in Section~\ref{sec:error_analysis_post_edit}. Table~\ref{tab:ende-post-edit-agree} presents the same data as in Figure~\ref{fig:ende_post_edit_agreement} for English-German, and Table~\ref{tab:zhen-post-edit-agree} presents the same for Chinese-English.   

\begin{table*}[!htp]
    \centering
    \begin{adjustbox}{width=0.98\textwidth}
    \begin{tabular}{C{2cm}C{2cm}C{2cm}C{2cm}C{2cm}C{2cm}C{2cm}}
    \toprule
    \textbf{Initial Translation} & \textbf{Total Seg}  & \textbf{HumanPE} & \textbf{LLMRefine} & \textbf{HumanPE \& LLMRefine} & \textbf{Human Only} & \textbf{LLMRefine Only}\\
    \midrule
    HumanRef & 460 & 163 (35.4\%) & 235 (51.1\%) & 110 (23.9\%) & 53 (11.5\%) & 125 (27.2\%)\\
    BestWMT & 460 & 314(68.3\%) & 281 (61.1\%) & 225 (48.9\%) & 89 (19.3\%) & 56 (12.2\%)\\
    MidWMT & 460 & 400 (87.0\%) & 324 (70.4\%) & 310 (67.4\%) & 90 (19.6\%) & 14 (3\%)\\
    \bottomrule
    \end{tabular}
    \end{adjustbox}
    \caption{Numerical breakdown of the agreement between \textsc{HumanPE} and \textsc{LLMRefine} in identifying segments requiring post-editing in English-German}
    \label{tab:ende-post-edit-agree}
\end{table*}

\begin{table*}[!htp]
    \centering
    \begin{adjustbox}{width=0.98\textwidth}
    \begin{tabular}{C{2cm}C{2cm}C{2.1cm}C{2cm}C{2cm}C{2cm}C{2cm}}
    \toprule
    \textbf{Initial Translation} & \textbf{Total Seg}  & \textbf{HumanPE} & \textbf{LLMRefine} & \textbf{HumanPE \& LLMRefine} & \textbf{Human Only} & \textbf{LLMRefine Only}\\
    \midrule
    HumanRef & 1175 & 558 (47.5\%) & 225 (19.1\%) & 161 (13.7\%) & 397 (33.8\%) & 64 (5.4\%)\\
    BestWMT & 1175 & 830 (70.6\%) & 133 (11.3\%) & 123 (10.5\%) & 707 (60.2\%) & 10 (0.9\%)\\
    MidWMT & 1175 & 1006 (85.6\%) & 408 (34.7\%) & 399 (34.0\%) & 607 (51.7\%) & 9 (0.8\%)\\
    \bottomrule
    \end{tabular}
    \end{adjustbox}
    \caption{Numerical breakdown of the agreement between HumanPE and LLMRefine in identifying segments requiring post-editing in Chinese-English}
    \label{tab:zhen-post-edit-agree}
\end{table*}





\end{document}